\pdfoutput=1
\documentclass[11pt,a4paper]{article}
\usepackage[hyperref]{acl2018}
\usepackage{times}
\usepackage{latexsym}
\usepackage{multicol}
\usepackage{booktabs}
\usepackage{url}
\usepackage{graphicx}
\usepackage[utf8]{inputenc}

\aclfinalcopy 

\title{Predicting Concreteness and Imageability of Words \\Within and Across Languages via Word Embeddings}

\author{Nikola Ljubešić \\
  Dept. of Knowledge Technologies \\
  Jožef Stefan Institute \\
  Jamova cesta 39, SI-1000 Ljubljana \\
  {\tt nikola.ljubesic@ijs.si} \\\And
  Darja Fišer \\
  Dept. of Translation, Faculty of Arts \\
  University of Ljubljana \\
  Aškerčeva 2, SI-1000 Ljubljana \\
  {\tt darja.fiser@ff.uni-lj.si}
  \\\AND
  Anita Peti-Stantić\\
  Faculty of Humanities and Social Sciences\\
  University of Zagreb\\
  Ivana Lučića 3, HR-10000 Zagreb\\
  {\tt anita.peti-stantic@ffzg.hr}
  }

\date{}

\begin{document}
\sloppy
\maketitle
\begin{abstract}
The notions of concreteness and imageability, traditionally important in psycholinguistics, are gaining significance in semantic-oriented natural language processing tasks. In this paper we investigate the predictability of these two concepts via supervised learning, using word embeddings as explanatory variables. We perform predictions both within and across languages by exploiting collections of cross-lingual embeddings aligned to a single vector space. We show that the notions of concreteness and imageability are highly predictable both within and across languages, with a moderate loss of up to 20\% in correlation when predicting across languages. We further show that the cross-lingual transfer via word embeddings is more efficient than the simple transfer via bilingual dictionaries.
\end{abstract}

\section{Introduction}

Concreteness and imageability are very important notions in psycholinguistic research, building on the theory of the double, verbal and non-verbal, modality of representation of concrete words in the mental lexicon, contrasted to single verbal representation of abstract words \cite{paivio1975coding,paivio2010dual}. 
Although often correlated with concreteness, imageability is not a redundant property. While most abstract things are hard to visualize, some call up images, e.g., \emph{torture} calls up an emotional and even visual image. There are concrete things that are hard to visualize too, for example, \emph{abbey} is harder to visualize than \emph{banana} \cite{Tsvetkov2014}.


Both notions have proven to be useful in computational linguistics as well.
\citet{Turney:2011:LMS:2145432.2145511} present a supervised model that exploits concreteness to correctly classify 79\% of adjective-noun pairs as having literal or non-literal meaning. \citet{Tsvetkov2014} exploit both the notions of concreteness and imageability to perform metaphor detection on subject-verb-object and adjective-noun relations, correctly classifying 82\% and 86\% instances, respectively.

The aim of this paper is to investigate the predictability of concreteness and imageability within a language, as well as across languages, by exploiting cross-lingual word embeddings as our available signal.

\section{Related Work}

While much work has been done on exploiting word embeddings in expanding sentiment lexicons \cite{tang2014building,amir2015inesc,HamiltonCLJ16}, there is little work on predicting other lexical variables, concreteness and imageability included.

\citet{Tsvetkov2014}
 performed metaphor detection, using, among others, concreteness and imageability as their features. To propagate these features, obtained from the MRC psycholinguistic database \cite{Wilson1988} to the entire lexicon, they used a supervised learning algorithm on vector space representations, where each vector element represented a feature. Performance of these classifiers was 0.94 for concreteness and 0.85 for imageability. They also applied the concreteness and imageability features to other languages by projecting features with bilingual dictionaries.

\citet{10.1007/978-3-642-37210-0_12} extended imageability scores to the whole lexicon by using the MRC imageability scores and hyponym and hyperonym links from WordNet.

\citet{DBLP:journals/corr/RotheES16} trained an orthogonal transformation to reorder word embedding dimensions into one-dimensional ultradense subspaces, the output thereby being a lexicon. They trained the transformations for sentiment, concreteness and frequency. For obtaining training data for concreteness, they used the BWK database \cite{5774089}. They showed that concreteness and sentiment can be better extracted from embedding spaces than frequency, with a Kendall $\tau$ correlation coefficient of 0.623 for concreteness. 
\citet{rothe2016word} further exploited this method to perform operations over the extracted dimensions, such as given a concrete word like \emph{friend}, find the related, but abstract word \emph{friendship}.


\paragraph{Contributions} 

In this paper we perform a systematic investigation of transfer of two lexical notions, concreteness and imageability, (1) to the remainder of the lexicon not covered in an annotation campaign, and (2) to other languages.

While there were already successful transfers within a language based on word embeddings \cite{Tsvetkov2014,rothe2016word}, the only cross-lingual transfer was based on transfer via bilingual dictionaries \cite{Tsvetkov2014}. In this paper we compare the effectiveness of cross-lingual transfer via word embeddings and via bilingual dictionaries.

A byproduct of this research is a lexical resource in 77 languages containing per-word estimates for concreteness and imageability.



\section{Data}


\subsection{\label{sec:lexicons}Lexicons}

In our experiments we use two existing English and one Croatian lexicon with concreteness and imageability ratings.

For English we use the MRC database \cite{Wilson1988} (\textsc{mrc} onwards), consisting of 4,293 words with ratings for concreteness and imageability. The ratings range from 100 to 700 and were obtained by merging three different resources \cite{Wilson1988}.

We also use the BWK database consisting of 39,954 English words \cite{5774089} (\textsc{bwk} onwards) with concreteness ratings summarized through arithmetic mean and standard deviation. The ratings were collected in a crowdsourcing campaign in which each word was labeled by 20 annotators on a 1--5 scale.

For Croatian we use the MEGAHR database (\textsc{mega} onwards), consisting of 3,000 words, with concreteness and imageability ratings summarized through arithmetic mean and standard deviation. The ratings were collected in an annotation campaign among university students, with each word obtaining 30 annotations per variable on a 1--5 scale.






For performing cross-lingual transfer via a dictionary, we use data from a large popular online Croatian-English dictionary\footnote{\url{http://www.taktikanova.hr/eh/}} containing around 100 thousand entries.

\subsection{Embeddings}

For both in-language and cross-lingual experiments we use the aligned Facebook collection of embeddings\footnote{\url{https://github.com/facebookresearch/fastText/blob/master/pretrained-vectors.md}}, trained with fastText \cite{bojanowski2016enriching} on Wikipedia dumps, with embedding spaces aligned between languages with a linear transformation learned via SVD \cite{Smith17} on a bilingual dictionary of 500 out of the 1000 most frequent English words, obtained via the Google Translate API\footnote{\url{https://github.com/Babylonpartners/fastText_multilingual}}.

We also experimented with another cross-lingual embedding collection \cite{conneau2017word}, obtaining similar results and backing all our conclusions.
This is in line with recent work on comparing cross-lingual embedding models which suggests that the actual choice of monolingual and bilingual signal is more important for the final model performance than the actual underlying architecture \cite{levy2017,ruder2017}. Given that one of our goals is to transfer concreteness and imageability annotations to as many languages as possible, using cross-lingual word embeddings based on Wikipedia dumps and dictionaries obtained through a translation API is the most plausible option.


\section{Experiments}

\subsection{Setup}

We perform two sets of experiments: one within each language, and another across languages.

While in-language experiments are always based on supervised learning, in cross-lingual experiments we compare two transfer approaches: one based on a simple dictionary transfer, and another on supervised learning on the word embeddings in the source language, and performing predictions on word embeddings in the target language, with the two embedding spaces being aligned.

We perform our prediction experiments by training SVM regression models (\textsc{svr}) and deep feedforward neural networks (\textsc{ffn}) over standardized (zero mean, unit variance) embeddings and each specific response variable. We experiment with all available gold annotations as our response variables, namely both the arithmetic mean and standard deviation of concreteness and imageability.

We tuned the hyperparameters of each of the regressors on a subset of the Croatian, \textsc{mega} dataset in the case of the in-language experiments, and another subset of the \textsc{bwk} dataset for the cross-lingual experiments. Given that we perform the final experiments on the whole datasets, and that we have two additional English datasets at our disposal for the in-language experiments and three additional dataset pairs for the cross-lingual experiments, we consider our approach to be resistant to the overfitting of the hyperparameters going unnoticed.

While the \textsc{svr} proved to work well with the RBF kernel, the $C$ hyperparameter of $1.0$ and the $\gamma$ hyperparameter of $0.003$, the feedforward network obtained strong results with two fully-connected hidden layers, consisting of 128 and 32 units each and ReLU activation functions, with a dropout layer after each of the hidden layers, and an output layer with a linear activation function. We optimized for the mean squared error loss function and ran 50 epochs on each of the datasets, with a batch size of 32.

While we used the same regressor setup for the \textsc{svr} system for both the in-language and cross-lingual experiments, for the \textsc{ffn} system the dropout probability in the in-language experiments was 0.5, while in the cross-lingual setting the dropout probability was set to 0.8, obtaining thereby a more general model which transfers better to the other language.

We perform in-language experiments via 3-fold cross-validation, while we train models on our source language dataset and evaluate the models on our target language dataset for cross-lingual experiments. We evaluate each approach via the Spearman rank and Pearson linear correlation coefficients. In the paper we report the Spearman correlation coefficient only as the relationships across both metrics in all the experiments are identical. We perform our experiments with the \texttt{scikit-learn} \cite{scikit-learn} and \texttt{keras} \cite{chollet2015keras} toolkits.

\begin{table*}
\centering
\begin{tabular}{l|rrrrrrrr}
\toprule
dataset & \multicolumn{2}{c}{\textsc{mega}} & \multicolumn{2}{c}{\textsc{bwk}} & \multicolumn{2}{c}{\textsc{bwk.3k}} & \multicolumn{2}{c}{\textsc{mrc}} \\
\midrule
lang & \multicolumn{2}{c}{hr} & \multicolumn{2}{c}{en} & \multicolumn{2}{c}{en} & \multicolumn{2}{c}{en} \\
size & \multicolumn{2}{c}{2,682} & \multicolumn{2}{c}{22,797} & \multicolumn{2}{c}{3,000} & \multicolumn{2}{c}{4,061} \\
method & \textsc{svr} & \textsc{ffn} & \textsc{svr} & \textsc{ffn} & \textsc{svr} & \textsc{ffn} & \textsc{svr} & \textsc{ffn} \\
\midrule
\textsc{c.m} & \textbf{0.760} & 0.742 & \textbf{0.887} & 0.879 & \textbf{0.848} & 0.834 & \textbf{0.872} & 0.863\\
\textsc{c.std} & \textbf{0.265} & \textbf{0.274} & \textbf{0.484} & 0.461 & \textbf{0.376} & \textbf{0.364} & - & - \\
\textsc{i.m} & \textbf{0.645}  & 0.602 & - & -  & - & - & \textbf{0.803} & 0.787 \\
\textsc{i.std} & \textbf{0.439} & \textbf{0.415} & - & - & - & - & - & - \\
\bottomrule
\end{tabular}
\caption{\label{tab:inlang}Results of the in-language experiments on predicting mean (\textsc{.m}) and standard deviation (\textsc{.std}) of concreteness (\textsc{c}) and imageability (\textsc{i}), either using a support vector regressor (\textsc{svr}) or feed-forward network (\textsc{ffn}). Evaluation metric is the Spearman correlation coefficient.}
\end{table*}

\subsection{\label{sec:inlang}In-language Experiments
}

We start our experiments in the in-language setting, running cross-validation experiments over each of our three datasets on all available variables. The results of these experiments, with some basic information on the size of the datasets, are given in Table \ref{tab:inlang}. Aside from the three lexicons introduced in Section \ref{sec:lexicons}, we experiment with another lexicon, \textsc{bwk.3k}, which is a randomly downsampled version of the \textsc{bwk} lexicon to the size of the two remaining lexicons. We introduce this additional resource (1) to control for dataset size when comparing results on our different datasets and (2) to measure the impact of training data size by comparing the results on the two flavours of the \textsc{bwk} dataset.

The results in Table \ref{tab:inlang} show that the support vector regressor consistently performs better than the feedforward neural network at predicting almost all values, with relative error reduction lying between 7\% and 12\%. The bold results are statistically significantly better than the corresponding non-bold ones given the approximate randomization test \cite{doi:10.1080/00223980.1969.10543491} with $p<0.05$. Our assumption is that the stronger \textsc{ffn} model does not show a positive impact primarily due to the small size of the datasets and the simplicity of the modeling problem.

We can further observe that the arithmetic mean is much easier to predict than standard deviation on both variables in all the datasets. This can be explained by the fact that standard deviation on the two phenomena can partially be explained with the level of ambiguity of a specific word, and this type of information is at least not directly available in context-based word embeddings.

Furthermore, imageability seems to be consistently slightly harder to predict than concreteness. Our initial assumption regarding this difference was that imageability is a more vague notion for human subjects, and therefore their responses are more dispersed, adding to the complexity of the prediction. However, analyzing standard deviations over concreteness and imageability showed that these are rather the same. We leave this open question for future research.

When comparing the results on predicting mean concreteness on the full \textsc{bwk} and the trimmed \textsc{bwk.3k} datasets, we see a significant improvement of the predictions of the on the larger dataset, showing that having 10 times more data for learning can produce significant improvements in the prediction quality.

\subsection{Cross-lingual Experiments}

\begin{table*}
\centering
\scalebox{0.95}{\begin{tabular}{l|rrr|rrr|rrr|rrr}
\toprule
source & \multicolumn{3}{c|}{\textsc{mega} (hr)} & \multicolumn{3}{c|}{\textsc{bwk} (en)} & \multicolumn{3}{c|}{\textsc{mega} (hr)} & \multicolumn{3}{c}{\textsc{mrc} (en)} \\
target & \multicolumn{3}{c|}{\textsc{bwk} (en)} & \multicolumn{3}{c|}{\textsc{mega} (hr)} & \multicolumn{3}{c|}{\textsc{mrc} (en)} & \multicolumn{3}{c}{\textsc{mega} (hr)} \\
 & \textsc{svr} & \textsc{ffn} & \textsc{dic} & \textsc{svr} & \textsc{ffn} & \textsc{dic} & \textsc{svr} & \textsc{ffn} & \textsc{dic} & \textsc{svr} & \textsc{ffn} & \textsc{dic} \\
\midrule
\textsc{c.m} & \textbf{0.791} & \textbf{0.793} & 0.728 & \textbf{0.724} & \textbf{0.719} & 0.641 & \textbf{0.797} & \textbf{0.794} & 0.611 & \textbf{0.651} & \textbf{0.644} & 0.638 \\
\textsc{c.std} & 0.178 & 0.141 & 0.224 & 0.185 & 0.145 & 0.137 & - & - & - & - \\
\textsc{i.m} & - & - & - & - & - & - & \textbf{0.694} & \textbf{0.683} & 0.523 & \textbf{0.548} & 0.531 & 0.503\\
\bottomrule
\end{tabular}}
\caption{\label{tab:crossling}Results of the cross-lingual experiments, either using supervised learning (\textsc{svr}, \textsc{ffn}), or simple dictionary lookup (\textsc{dic}). Evaluation metric is the Spearman correlation coefficient. Results in bold are best results per problem with no statistically significant difference.}
\end{table*}

In cross-lingual experiments we compare our two approaches to cross-lingual transfer: dictionary lookup (\textsc{dic} onwards) and supervised learning on aligned word embedding spaces via the two methods introduced in Section \ref{sec:inlang}, \textsc{svr} and \textsc{ffn}.

The \textsc{dic} method simply looks up for each word in the source language resource all possible translations to the target language and directly transfers the concreteness and imageability ratings to the target language words. In case of collisions in the target language (two source language words being translated to the same word in the target language), we perform averaging over the transfered ratings. In our experiments, the arithmetic mean showed to be a better averaging method than the median, we therefore report the results on that averaging method.

The \textsc{svr} and \textsc{ffn} methods use supervised learning in a very similar fashion to the in-language experiments described in Section \ref{sec:inlang}. We train a supervised regression model on the whole source language dataset, using word embedding dimensions as features and the variable of choice as our target. We obtain estimates of our variable of choice in the target language by applying the source-language model on the target-language word embeddings since the two embedding spaces are aligned.

For both approaches we compare the target-language estimates with the gold data available from our lexicons.

We present the results of the cross-lingual experiments in Table \ref{tab:crossling}. Our first observation is that, while in the in-language setting the \textsc{svr} method has regularly outperformed the \textsc{ffn} method, in the cross-lingual setting this is not the case any more, with \textsc{svr} and \textsc{ffn} obtaining very similar results, in five out of six cases in the range of no statistically significant difference. Our explanation for the loss of the positive impact in using the weaker, support vector regression model, is that with the noisy alignment of the two embedding spaces the prediction problem became harder, now both models performing similarly. While the strong point of \textsc{svr} is that it performs very well on small datasets, the strong point of the \textsc{ffn} method is that it generalizes better.

That higher generalization is beneficial in case of the cross-lingual problem is observable in the difference in the hyperparameter tuning results on the \textsc{ffn} method, where in the in-language setting the optimal dropout was 0.5, while in the cross-lingual setting it is 0.8.

Our second observation is that all the predicted ratings suffer in the cross-lingual setting, when compared to the in-language results presented in Table \ref{tab:inlang}, observing for the \textsc{svr} method a drop of around 5 to 15\%. While standard deviation was already poorly predicted in the in-language setting, in the cross-lingual setting it drops even further to a non-useful level, below 0.2. This is the reason why we do not calculate statistical significance of the differences in these results and do not include their estimates in our final 77-languages-strong resource. In the final cross-lingual resource we include only the mean of concreteness and imageability, the notions for which we have obtained strong correlation in our cross-lingual experiments.


Finally, when comparing the cross-lingual transfer via embeddings (\textsc{svr} and \textsc{ffn}) and via a dictionary (\textsc{dic}), the learning-on-embeddings approach outperforms the dictionary method in each instance, with the relative loss in correlation when moving from the embedding to the dictionary approach of 5\% to 25\%.

\subsection{Regressor Coefficient Analysis}

Our final analysis concerns the question of how many of the embedding dimensions are crucial for our regressors to predict the notions of concreteness and imageability. We consider two potential scenarios: (1) each of the notions are encoded in one or a few of the embedding dimensions and (2) the notions are encoded in many embedding dimensions.

The analysis is performed by calculating the cumulative distribution of absolute and normalized (sum to 1), reversely sorted coefficients of the SVM regressor with a linear kernel. For both phenomena, concreteness and imageability, the distributions show that the predictions are based on a significant number of embedding dimensions. Namely, while 80 most informative dimensions cover 50\% of the coefficients' mass, half of the dimensions (150) cover 80\% of that mass. This shows for the second scenario -- concreteness and imageability are encoded in a significant number of embedding dimensions -- to be true.



\section{Conclusion}

In this paper we have shown that concreteness and imageability ratings can be successfully transfered both to non-covered portions of the lexicon and to other languages via (cross-lingual) word embeddings.

With the in-language experiments we have shown that the arithmetic mean of both notions is much easier to predict than their standard deviation, the latter probably encoding word ambiguity, type of information not directly present in word embeddings. 

Our experiments across languages have shown that the loss in comparison to in-language experiments on predicting the means of both concreteness and imageability are around 15\%, a reasonable price to pay given the applicability of the method to all of the 77 languages present in the word embedding collection. The predictions of concreteness and imageabililty obtained in the 77 languages are available at \url{http://hdl.handle.net/11356/1187}.\footnote{Ongoing developments are stored at \url{https://github.com/clarinsi/megahr-crossling/}.}


Comparing the two methods of transfer -- dictionary vs. cross-lingual embeddings, shows regularly better (5\%--15\%) results of the latter, proving once more the usefulness of word embeddings, especially in the currently expanding cross-lingual setup.

Finally, while the stronger deep neural model shows worse results than the support vector regressor in the in-language setting, mainly because of the small size of the training datasets, in the cross-lingual setting they both show identical performance due to the problem becoming harder given the noise from the embedding alignment process.

\section*{Acknowledgements}

The work described in this paper has been funded by the Croatian National Foundation project HRZZ-IP-2016-06-1210, the  Slovenian  Research  Agency project ARRS J7-8280, and by the Slovenian research infrastructure CLARIN.SI.


\bibliographystyle{acl_natbib}
\bibliography{acl2018}

\end{document}